\documentclass[10pt,twocolumn,letterpaper]{article}

\usepackage{iccv}
\usepackage{times}
\usepackage{epsfig}
\usepackage{graphicx}
\usepackage{amsmath}
\usepackage{amssymb}
\usepackage{booktabs}
\usepackage{marvosym}
\usepackage{algpseudocode}
\usepackage{algorithm}
\usepackage{subcaption}
\usepackage{xcolor}
\usepackage{colortbl}
\usepackage{multirow}

\usepackage[pagebackref=true,breaklinks=true,letterpaper=true,colorlinks,bookmarks=false]{hyperref}

\iccvfinalcopy % *** Uncomment this line for the final submission

\ificcvfinal\fi

\begin{document}

%%%%%%%%% TITLE
\title{Pseudo-label Correction and Learning For Semi-Supervised Object Detection}

\author{Yulin He$^1$\thanks{Equal contribution.}    ~~~~~~~
Wei Chen$^1$\footnotemark[\value{footnote}] \thanks{Corresponding author.}  ~~~~~~~
Ke Liang$^1$ ~~~~~~~ \\
Yusong Tan$^1$ ~~~~~~~
Zhengfa Liang$^2$ ~~~~~~~
Yulan Guo$^1$ \\
$^1$ National University of Defense Technology \\
$^2$ Defense Innovation Institute, Chinese Academy of Military Science \\
{\tt\small \{heyulin, chenwei, liangke, ystan, liangzhengfa10, yulan.guo\}@nudt.edu.cn}}

% For a paper whose authors are all at the same institution,
% omit the following lines up until the closing ``}''.
% Additional authors and addresses can be added with ``\and'',
% just like the second author.
% To save space, use either the email address or home page, not both

\maketitle
% Remove page # from the first page of camera-ready.
\ificcvfinal\thispagestyle{empty}\fi

%%%%%%%%% ABSTRACT
\begin{abstract}
  Pseudo-Labeling has emerged as a simple yet effective technique for semi-supervised object detection (SSOD). 
  However, the inevitable noise problem in pseudo-labels significantly degrades the performance of SSOD methods. 
  Recent advances effectively alleviate the classification noise in SSOD, while the localization noise which is a non-negligible part of SSOD is not well-addressed. 
  In this paper, we analyse the localization noise from the generation and learning phases, and propose two strategies, namely pseudo-label correction and noise-unaware learning. 
  For pseudo-label correction, we introduce a multi-round refining method and a multi-vote weighting method.
  The former iteratively refines the pseudo boxes to improve the stability of predictions, while the latter smoothly self-corrects pseudo boxes by weighing the scores of surrounding jittered boxes. 
  For noise-unaware learning, we introduce a loss weight function that is negatively correlated with the Intersection over Union (IoU) in the regression task, which pulls the predicted boxes closer to the object and improves localization accuracy. 
  Our proposed method, Pseudo-label Correction and Learning (PCL), is extensively evaluated on the MS COCO and PASCAL VOC benchmarks.
  On MS COCO, PCL outperforms the supervised baseline by 12.16, 12.11, and 9.57 mAP and the recent SOTA (SoftTeacher) by 3.90, 2.54, and 2.43 mAP under 1\%, 5\%, and 10\% labeling ratios, respectively.
  On PASCAL VOC, PCL improves the supervised baseline by 5.64 mAP and the recent SOTA (Unbiased Teacherv2) by 1.04 mAP on AP$^{50}$.
\end{abstract}

\begin{figure}[t]
   \begin{center}
   \includegraphics[width=0.95\linewidth]{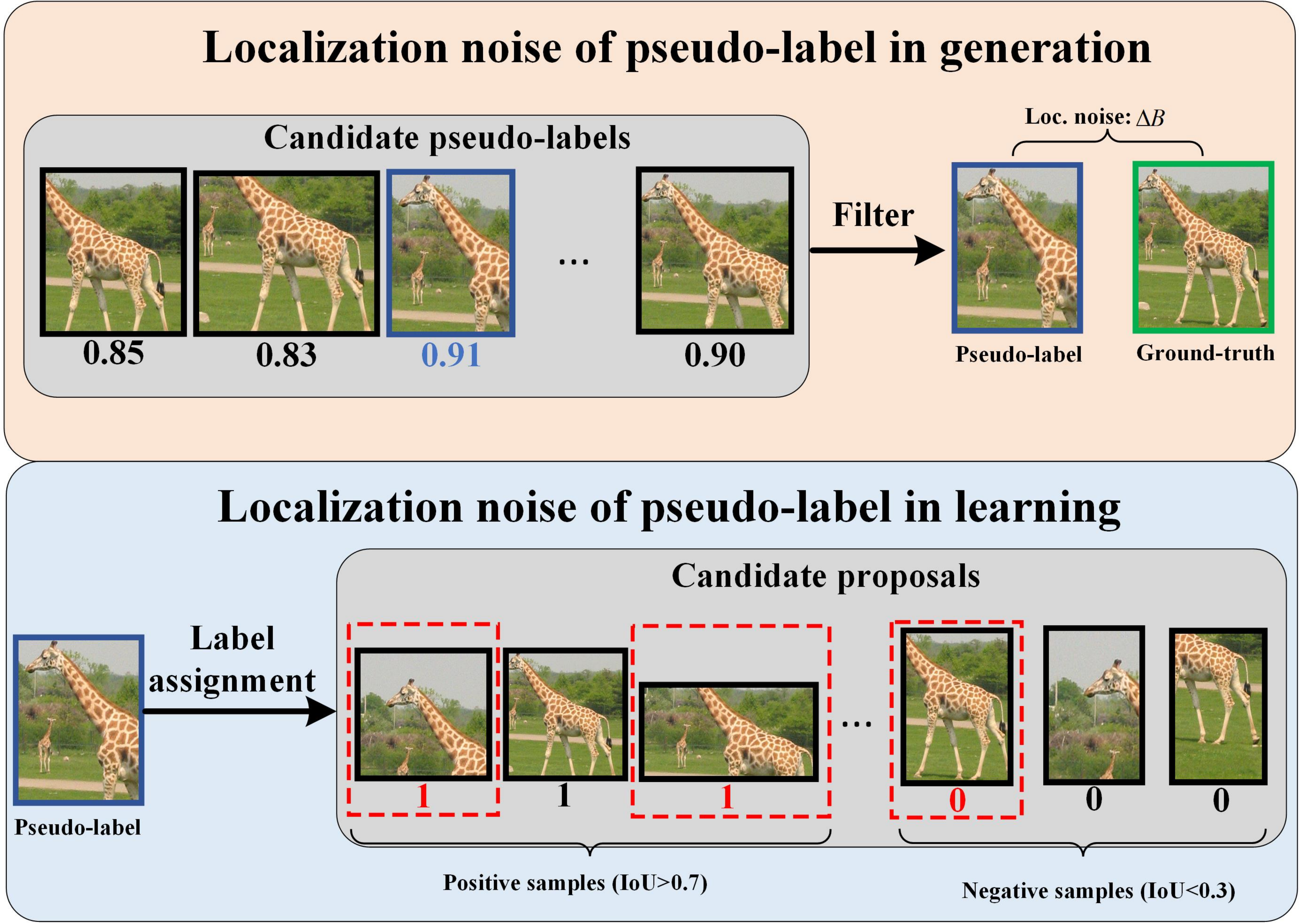}
   \end{center}
   \vspace{-3mm}
   \caption{
           Localization noise in pseudo-labels comes from two phases: generation and learning.
           In the generation phase, some candidate pseudo-labels with inaccurate localization receive high scores (i.e., the box drawn in blue), leading to the positional mismatch between pseudo-labels and ground-truth.
           In the learning phase, the inaccurate pseudo-labels cause a serious noisy problem in label assignment.
           Specifically, some candidate proposals are incorrectly assigned with wrong training targets (i.e., the dashed boxes drawn in red), which confuses the model in judging which candidate proposal should give a high score.
           As these two phases are conducted alternately during training, the localization noise will be accumulated, making training more challenging.}
    \label{fig:bar}
    \vspace{-3mm}
 \end{figure}

%%%%%%%%% BODY TEXT
\section{Introduction}
\label{sec:intro}

Benefiting from large-scale dataset and growing computational power, deep neural networks have led to great success in computer vision.
In the standard supervised learning protocol, the volume and quality of labels are extremely crucial to the performance.
However, collecting human annotations is labor-intensive and time-consuming, especially for instance-level annotations such as object detection.
Semi-supervised learning (SSL) is an effective remedy that utilizes a small portion of labeled data and abundant unlabeled data to achieve satisfactory performance.
SSL has achieved significant progress in classification tasks in recent years, with performance comparable to fully supervised learning ~\cite{fixmatch,simmatch}.
However, semi-supervised object detection (SSOD), which is more challenging than semi-supervised classification, has not been adequately explored.

Recent state-of-the-art SSOD methods \cite{stac,softteacher,interactive,instant} are based on pseudo-labeling with a mean-teacher structure~\cite{meanteacher}.
The student model is supervised using the pseudo-labels generated by the teacher model, so the quality of pseudo-labels directly impacts the performance of SSOD.
Nevertheless, the noise problem of pseudo-labels significantly degrades the performance of SSOD.
This noise consists of classification noise (i.e., misclassification) and localization noise (i.e., localization inaccuracy).
The former one is treated as a confirmation bias issue in SSL, many off-the-shelf methods~\cite{softteacher,scmt,labelmatch} are proposed to significantly reduce the classification errors.
However, the latter one, as a non-negligible component of pseudo-label noise, has not been well addressed.
As shown in Fig.~\ref{fig:bar}, localization noise impacts the training of SSOD from generation and learning phases.
Pseudo-labels with poor localization quality will lead to incorrect label assignment, which easily makes the model confused and severely degrades the detection performance.  
Therefore, \textit{(1) how to guarantee the localization quality of pseudo-labels?} and \textit{(2) how to learn to localize objects from noisy pseudo-labels?} are the concerns of this paper.

For the first concern, significant efforts~\cite{stac,unbiasedv1,softteacher} have been made to design appropriate selection strategies to filter out low-quality pseudo-labels, thereby ensuring the localization quality of pseudo-labels.
However, few have considered to directly revise the pseudo-labels in a self-correcting manner.
We found the inaccuracy of the pseudo-labels comes from two aspects, i.e., the prediction instability and the unreliability of NMS (Non-Maximum Suppression).
To address the above two issues, a pseudo-label correction method is proposed to self-correct pseudo-labels with two steps, i.e., multi-round refining and multi-vote weighting.
The primary objective of the multi-round refining step is to improve the stability of pseudo-labels.
To achieve this, we repeatedly feed the pseudo boxes into the teacher's RCNN layer for refinement.
As for multi-vote weighting step, we argue that the prediction boxes are interrelated, in contrast to NMS which considers them independently. 
Specifically, we introduce the box jitter function to the pseudo boxes with a particular variance. 
Then, we correct the pseudo boxes by assigning weights to the scores of adjacent jittered boxes.
Pseudo-label correction is a heuristic algorithm that uses the output score as an indicator to search for an optimized box locally within the feature map space.

For the second concern, the research of it is still in the early stage.
Some existing approaches include Unbiased Teacher~\cite{unbiasedv1}, which achieves better performance without learning the pseudo-label localization;
Dense Teacher~ \cite{dense}, which utilizes the teacher model's predictions as soft-labels to guide the localization learning of the student model;
and SCMT\cite{scmt}, which introduces a localization loss weight function proportional to the IoU, enhancing the learning of confident pseudo boxes. 
Despite these efforts, there is no conclusive answer on how to effectively learn the regression task from pseudo-labels.
In our framework, we compared the aforementioned methods and found that designing soft weights that are negatively correlated with the IoU value leads to the best results.
These results suggest that pseudo boxes with high confidence may not be as accurate as human annotations, they can also guide a relatively accurate optimization direction. 

The main contributions of this work are as follows:
\begin{itemize}
    \item We propose a pseudo-label correction method to improve the stability and localization accuracy of pseudo-labels, thus providing a more accurate supervised signal to the student model.
    \item We provide an in-depth analysis of label and weight designs in the unsupervised regression loss, and then propose a novel soft-weight function that can effectively improve the localization accuracy.  
    \item We validate the performance and transferability of our method on three recent SSOD algorithms and two popular datasets, and the results show consistent and significant accuracy improvements.
\end{itemize}
%------------------------------------------------------------------------
\begin{figure*}[t]
  \begin{center}
  \includegraphics[width=0.95\linewidth]{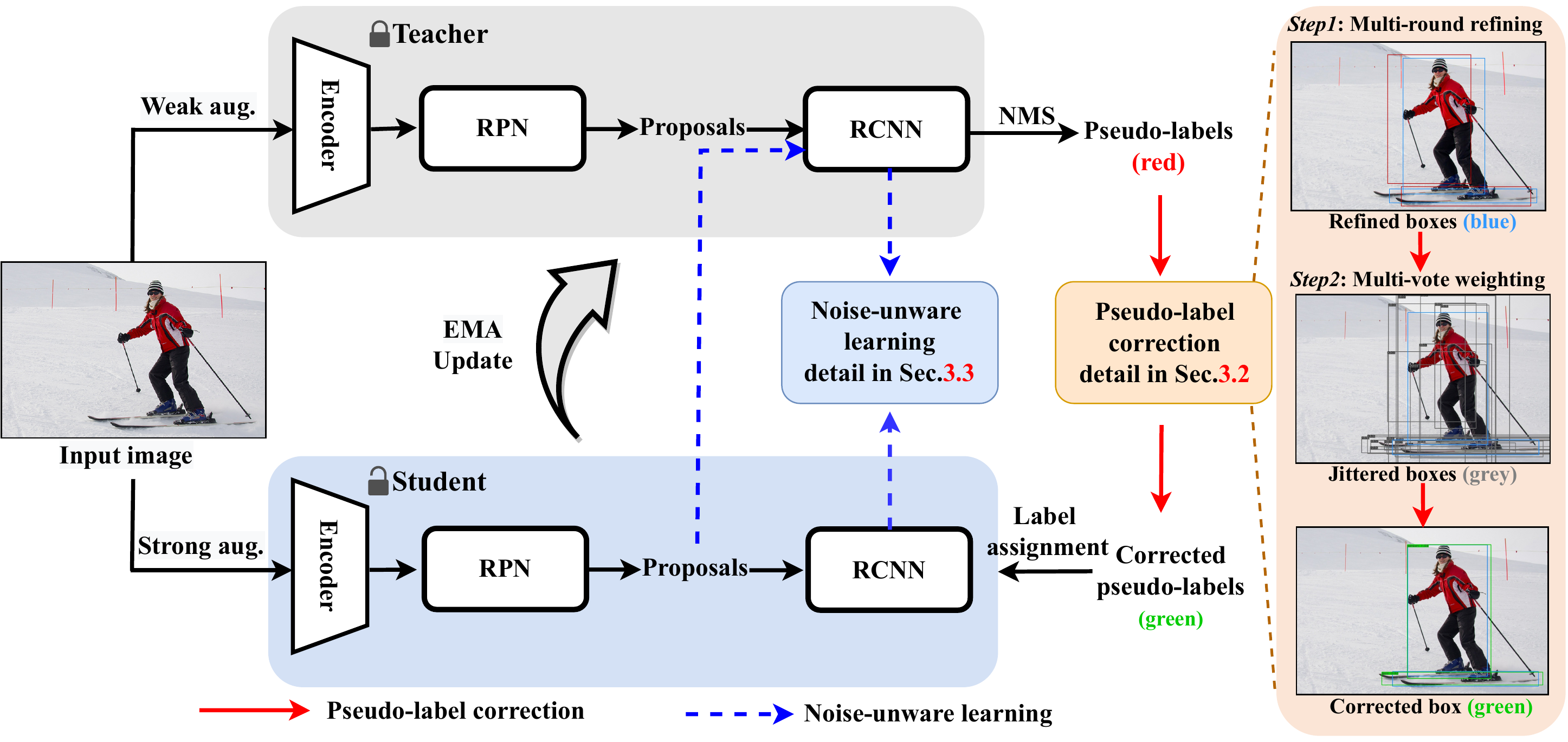}
  \vspace{-3mm}
  \end{center}
  \caption{Overview of our proposed method. 
          Firstly, weakly augmented images are fed into the teacher model to generate pseudo-labels. 
          To ensure the accuracy of the pseudo-labels, we apply the pseudo-label correction method to self-correct them. 
          Then, strongly augmented images are input to the student model to generate proposals, which are further processed by the teacher's RCNN layer to obtain soft labels and soft weights, as indicated by the dotted lines. 
          Finally, we employ the noise-unaware learning strategy to compute unsupervised loss.
          % For the best view, we increase the variance in box jitter.
          }
  \label{fig:framework}
  \vspace{-3mm}
\end{figure*}

\section{Related Works}
\textbf{Semi-supervised learning for image classification:} 
The majority of SSL methods for image classification can be roughly categorized into two types: consistency regularization and pseudo-labeling. 
Consistency regularization methods~\cite{bachman2014learning,sajjadi2016regularization} rely on the assumption that no matter what perturbations are applied, the model should keep the consistent output for the same unlabeled data. 
Data augmentation~\cite{bachman2014learning}, model perturbation~\cite{laine2016temporal}, and adversarial training~\cite{miyato2018virtual} are the main strategies to introduce perturbations. 
Additionally, a mean teacher~\cite{meanteacher} structure is proposed, which ensembles time-series models through exponential mean average (EMA) to form a teacher model, and then guides the student model to maintain consistency during training.
Pseudo-labeling methods~\cite{lee2013pseudo,fixmatch} employ weak data augmentation to annotate pseudo-labels for unlabeled data.
Due to their general and task-agnostic advantages, they have been widely used in SSL.
However, they tend to underperform due to the noise of pseudo-labels.  
\cite{mixmatch, remixmatch, zheng2022simmatch} explore the consistency between weak and strong data augmentation through soft pseudo-labels, mitigating the harmful impacts of noise.

\textbf{Semi-supervised object detection:} 
Inspired by SSL, pseudo-labeling and consistency regularization are the mainstream methods of SSOD.
In \cite{consistency,interpolation}, flip consistency and interpolation consistency are used to perform instance-level consistency for SSOD.
STAC \cite{stac} firstly applies pseudo-labels performed on weak and strong data augmentation into SSOD.
Following STAC \cite{stac}, \cite{unbiasedv1,instant,softteacher} further develope pseudo-labeling and achieve a significant performance improvement.
However, the noise of pseudo-labels has always been the crux of pseudo-labeling methods.
Generally, the noise in pseudo-labels consists of two parts, including classification noise and localization noise.

\textbf{Classification noise in semi-supervised object detection:} 
Classification noise has already attracted a lot of attention in the SSOD community.
Humble Teacher~\cite{humble} exploits KL divergence to narrow classification output distributions of the teacher and the student model. 
LabelMatch~\cite{labelmatch} utilizes soft-labels generated by the teacher model to guide the learning of the student model.
SoftTeacher~\cite{softteacher} adds soft-weights to negative pseudo-labels to alleviate missing detection problem.
SCMT~\cite{scmt} proposes to dynamically re-weight the unsupervised loss of each proposal from both RPN and RCNN parts. 
After extensive research, adopting soft labeling or soft weighting techniques in SSOD can greatly alleviate the impact of classification noise.

\textbf{Localization noise in semi-supervised object detection:} 
While, the research of the localization noise is still in the early stage.
Many existing works guarantee the quality of pseudo-labels by designing pseudo-label selection strategies.
For instance, STAC \cite{stac} and SoftTeacher~\cite{softteacher} simply set a high filter threshold.
LabelMatch~\cite{labelmatch} proposes an adaptive thresholding strategy, which considers the distribution of labeled data and sets different thresholds for different categories.
RPL~\cite{rethinking} proposes to add an extra certainty-aware task to estimate regression quality of pseudo-labels.
Unbiased Teacherv2~\cite{unbiasedv2} proposes to predict uncertainties on the regression branch to select pseudo-labels for boundary prediction.
However, these methods only focus on selecting pseudo-labels rather than correcting them during generation.
In contrast, we propose to self-correct the pseudo-labels by enhancing their stability and relatedness.
We argue that our proposed pseudo-label correction strategy is orthogonal to the above works and can work collaboratively with them.
Another branch of research is on learning localization knowledge from noisy pseudo-labels, with various strategies proposed, such as hard labels \cite{stac, softteacher}, soft labels \cite{dense}, and soft weights \cite{scmt, pseco}.
In our framework, we compared these three types of techniques, and proposed a novel learning strategy that can tightly pull the prediction boxes around the objects.

\section{Methodology}
\label{sec:method}

The pipeline of our proposed method is illustrated in Fig.~\ref{fig:framework}.
In Sec.~\ref{sec:Preliminary}, we present the preliminary and introduce the baseline of our method.
Then, we propose a pseudo-label correction strategy in Sec.~\ref{sec:splc}, which aims to self-correct pseudo-labels.
However, noise still exists after correction and negatively affects performance.
To address this, we propose a noise-unaware learning strategy in Sec.~\ref{sec:spll} to combat the noise in pseudo-label learning.

\subsection{Preliminary}
\label{sec:Preliminary}

\textbf{Problem Definition:} In SSOD, a set of labeled data $D_s = \{x_i^s, y_i^s\}_{i=1}^{N_s}$ and a set of unlabeled data $D_u = \{x_i^u \}_{i=1}^{N_u}$ are given.
$x_i$ and $y_i$ are the image and annotation of $i$-th sample, where $y_i$ contains the categories and locations of the objects in $i$-th sample.
$N_s$ and $N_u$ denote the numbers of $D_s$ and $D_u$. 
The goal of SSOD is to leverage massive unlabeled data to boost the performance of object detection.

\textbf{SSOD with Mean Teacher Framework:} 
Our SSOD baseline is built upon the mean teacher framework with pseudo-label learning.

Mean teacher framework consists of a teacher model and a student model, which are optimized through a mutual learning mechanism.
The weights of the teacher model are updated by the EMA (Exponential Moving Average) weights of the student model, while the student model exploits gradient backpropagation to update its weights.
The teacher model predicts the pseudo-labels of unlabeled data, and the student model is trained by both labeled data and unlabeled data.
Therefore, the loss of SSOD can be summarized as a supervised loss $\mathcal{L}_s$ on labeled data and an unsupervised loss $\mathcal{L}_u$ on unlabeled data.
{
\begin{equation}\
\label{eq:loss}
\begin{aligned}
  \mathcal{L}= \frac{1}{N_s} \sum_{i=1}^{N_s} \mathcal{L}_s(x^s, y^s)+  \frac{1}{N_u} \sum_{i=1}^{N_u} \alpha \mathcal{L}_u(x^u, y^u),
\end{aligned}
\end{equation}}where $N_s$ and $N_u$ are the number of labeled data and unlabeled data, respectively. 
$y^u$ is the pseudo-labels consisting of classification and regression parts.
$\alpha$ controls the contribution of $\mathcal{L}_u$.

\begin{figure*}[t]
  \begin{center}
  \includegraphics[width=1.0\linewidth]{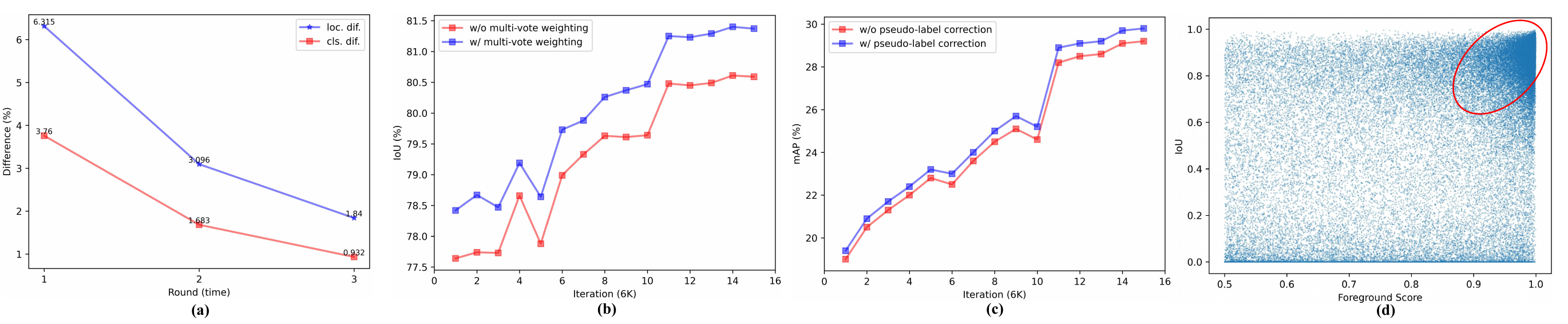}
  \end{center}
  \caption{We randomly sample 10k unlabeled images to conduct experiments based on the model trained at the 10\% labeled data setting.
          (a) denotes the changes of multi-round refining, consisting of IoU and score.
          (b) and (c) show the IoU and mAP between pseudo-labels (confidence $\geq$ 0.7 ) and the ground-truth.
          (d) describes the relations between the IoU with the ground-truth and the foreground scores. 
          Each scatter represents one pseudo-label.
          The red circle indicates the dense place.}
  \label{fig:zhexian1}
\end{figure*}

\begin{algorithm}[t]
  \caption{Pseudo-label correction}
  \label{algorithm}
  \renewcommand{\algorithmicrequire}{\textbf{Input:}} % Use Input in the format of Algorithm
  \renewcommand{\algorithmicensure}{\textbf{Output:}} 
  \begin{algorithmic}[1]
  \Require
  RCNN layer $\mathcal{M}$;
  Feature map $F$; 
  Initial pseudo boxes ${B^0}$; 
  Number of the refined rounds $n_r$; 
  Number and variance of jittered boxes $n_j, \sigma$
  \Ensure
  Corrected boxes ${B_c}$ and scores ${L_c}$
  \For{$r=1$ to $n_r$} (\textbf{multi-round refining})
  \State  \textcolor{gray}{\textit{// Refine boxes and scores}} 
  \State $B^r, S^r = \mathcal{M}(F, B^{r-1})$
  \EndFor 
  \State $B_c$ = \{\}
  \For{$b$ in $B^{n_r}$} (\textbf{multi-vote weighting})
  \State  \textcolor{gray}{\textit{// Generate jittered boxes based on $b$}} 
  \State $B_j = \mathcal{F}_{j} (b, n_j, \sigma)$
  \State  \textcolor{gray}{\textit{// Calculate prediction boxes and scores of $B_j$}} 
  \State $\hat{B_j}, S_j = \mathcal{M}(F, B_j)$
  \State  \textcolor{gray}{\textit{// Correct pseudo boxes by weighted summation}} 
  \State $b_c = \sum_{i=1}^{n_j}\hat{B_j^i} \times \frac{S_j^i}{\sum_{i=1}^{n_j}S_j^i}$;
  \State $B_c$ $\gets$ \textit{Append}($b_c$)
  \EndFor
  \State  $S_c = S^{n_r}$; \\
  \Return $B_c$, $S_c$
  \end{algorithmic}
\end{algorithm}

\subsection{Pseudo-label correction}
\label{sec:splc}

Pseudo-label correction can be regarded as a heuristic algorithm that improves the localization accuracy of pseudo boxes using the model's output as a guide. 
It involves two main steps: multi-round refining and multi-vote weighting, as detailed in Algo.~\ref{algorithm}.

\textbf{Multi-round refining:} 
Before we present the detail of multi-round refining, we first introduce two metircs to meansure the differences between multiple rounds  in both classification and localization:
{
\begin{equation}
\begin{aligned}
 \mathcal{D}_{cls}^{r} &= \frac{\sum_{i=1}^{n_{pse}} |S_{i}^{r} - S_{i}^{r-1}|}{n_{pro}},
\end{aligned}
\end{equation}}
{
\begin{equation}
\begin{aligned}
 \mathcal{D}_{loc}^{r} &= 1 - \frac{\sum_{i=1}^{n_{pse}} IoU(B_{i}^{r}, B_{i}^{r-1})}{n_{pro}},
\end{aligned}
\end{equation}}where $n_{pse}$ is the number of pseudo-labels, $B^{r}$ and $S^{r}$ are the bounding boxes and scores of $r$-th round refinement.
$S^{0}$ and $B^{0}$ are the scores and boxes of initial pseudo-labels.
$\mathcal{D}_{cls}$ and $\mathcal{D}_{loc}$ are the classification and localization differences between two rounds, respectively.
The smaller the values of $\mathcal{D}_{cls}$ and $\mathcal{D}_{loc}$, the more stable the prediction.

Specifically, we alternately input prediction boxes and the basic feature map into the RCNN layer of the teacher model to obtain continuously refined boxes.
As the boxes are refined, the prediction scores change simultaneously.
As shown in Fig~\ref{fig:zhexian1} (a), after multi-round refining, the output of the teacher model can be more stable.
Furthermore, We observed that this process reaches saturation after several rounds of refinement, indicating that the model is confident and requires little correction.
The intuitive assumption behind multi-round refining is that \textbf{the smaller the output variance between two rounds, the more confident the model is towards the output, and thus the pseudo-labels are more accurate.}

\textbf{Multi-vote weighting:} NMS is a standard post-process operation used in object detection.
Its ``winner-takes-all'' strategy speeds up the post-processing and effectively suppresses overlapping false-positive bounding boxes. 
However, due to inaccurate pseudo boxes, some inaccurately localized boxes are assigned high scores, which makes it unreliable to rely on the score of individual boxes to determine the output box.
Therefore, we propose to exploit the scores of the surrounding boxes together to decide the position of the output box.

Specifically, we jitter the refined boxes multiple times with a certain variance, where one refined box corresponds to multiple jittered boxes. 
The box jitter function $\mathcal{F}_{j}$ is executed in the following steps: \\
(1) Sample jitter factors $\xi$ by a standard normal distribution with the variance of $\sigma_j$.
{
\begin{equation}
   \label{eq:jitter}
\begin{aligned}
  \xi \sim Sample(N(0, \sigma_j), n_j), \ \xi \in \mathbb{R}^{n_j \times 4}.
\end{aligned}
\end{equation}}
(2) Get jittered boxes by adding jittered offsets.
{
\begin{equation}
\begin{aligned}
  B_j =  (1 + \xi) \times Expand(b, n_j), 
\end{aligned}
\end{equation}}where $Expand$ denotes the function that expands $b$ from the size of $[{1 \times 4}]$ to $ [{n_j \times 4}]$. Then, The jittered boxes are fed into the RCNN layer again to obtain their prediction boxes and scores.
Finally, we normalize the prediction scores and use them as weights to sum the prediction boxes, getting the final self-corrected boxes.
This method transforms the localization of pseudo boxes from being solely determined by the box with the maximum score to being voted by the surrounding boxes.
As shown in Fig~\ref{fig:zhexian1} (b), after multi-vote weighting, the localization of output boxes can be more accurate.
The intuitive assumption behind it is that \textbf{inaccurate boxes will be corrected by surrounding boxes that are close to objects, as they are likely to have high weights.}

After applying the pseudo-label correction, the quality of pseudo-labels is improved, which can be demonstrated by the empirical study in Fig.~\ref{fig:zhexian1} (c).

\begin{table*}[t]
  \centering
  \renewcommand{\arraystretch}{1.3}
  \caption{
    The $mAP$ performance comparisons of the state-of-the-art SSOD methods on val2017 data set under COCO-standard. 
    }
  \label{table:mscoco}
  \scalebox{0.9}{
  \begin{tabular}{lccc}
  \toprule
  Method& 1 \% labeled data    &    5 \% labeled data    &    10 \% labeled data        \\ \hline
  Supervised& 12.20 $\pm$ 0.29      &    21.17 $\pm$ 0.17    &    26.90 $\pm$ 0.08      \\ \hline
  STAC~\cite{stac}    & 13.97 $\pm$ 0.35 & 24.38 $\pm$ 0.12 & 28.64 $\pm$ 0.21 \\
  Humble Teacher~\cite{humble} & 16.95 $\pm$ 0.35 & 27.70 $\pm$ 0.15 & 31.61 $\pm$ 0.28 \\ 
  ISMT ~\cite{interactive} & 18.88 $\pm$ 0.74 & 26.37 $\pm$ 0.24 & 30.53 $\pm$ 0.52 \\ 
  Unbiased Teacher~\cite{unbiasedv1}    & 20.75 $\pm$ 0.12      & 28.27 $\pm$ 0.11       & 31.50 $\pm$ 0.10 \\
  SoftTeacher~\cite{softteacher} & 20.46 $\pm$ 0.39       & 30.74 $\pm$ 0.08      & 34.04 $\pm$ 0.14 \\
  SCMT~\cite{scmt} & 23.09 $\pm$ 0.16  & 32.14 $\pm$ 0.06      & 35.42 $\pm$ 0.12 \\
  RPL \cite{rethinking} & 19.02 $\pm$ 0.25 & 28.40 $\pm$ 0.15 & 32.23 $\pm$ 0.14  \\  
  ACRST~\cite{ACRST} & \textbf{26.07 $\pm$ 0.46} & 31.35 $\pm$ 0.13 & 34.92 $\pm$ 0.22 \\
  Unbiased Teacher v2~\cite{unbiasedv2}    & 25.40 $\pm$ 0.36      & 31.85 $\pm$ 0.09       & 35.08 $\pm$ 0.02  \\
  LabelMatch~\cite{labelmatch}    & 25.81 $\pm$ 0.28      & 32.70 $\pm$ 0.18       & 35.49 $\pm$ 0.17  \\
  MA-GCP~\cite{MAGCP} & 21.30 $\pm$ 0.28 & 31.67 $\pm$ 0.16 & 35.02 $\pm$ 0.26 \\ 
  \hline
  PCL (Ours)   & 24.36 $\pm$ 0.12       & \textbf{33.28 $\pm$ 0.09}       & \textbf{36.47 $\pm$ 0.15 } \\
  \bottomrule
  \end{tabular}}
\end{table*}

\begin{table}[t]
   \renewcommand\arraystretch{1.3}
   \caption{The $mAP$ performance comparisons of the state-of-the-art SSOD methods on val2017 data set under COCO-additional. }
   \vspace{-3mm}
   \label{table:mscoco_additional}
   \begin{center}
   \scalebox{0.9}{
   \begin{tabular}{cc}
   \toprule
   Methods & $mAP$ \\ \hline
   Supervised &  37.63 \\
   CSD~\cite{consistency}  &     38.82 \\
   STAC~\cite{stac} &    39.21 \\
   Instant-Teaching~\cite{instant} &  40.20 \\
   Unbiased Teacher v1~\cite{unbiasedv1}  & 41.30 \\
   SoftTeacher~\cite{softteacher}  & 41.40 \\
   SCMT ~\cite{scmt}   & 42.56 \\
   PCL (Ours)   &  \textbf{44.33} \\
   \bottomrule 
   \end{tabular}}
   \end{center}
   \vspace{-3mm}
 \end{table}
 
 \begin{table}[t]
   \renewcommand\arraystretch{1.3}
   \caption{The $AP^{50}$ performance comparisons of the state-of-the-art SSOD methods under \textit{VOC}.}
   \vspace{-3mm}
   \label{table:voc}
   \begin{center}
   \scalebox{0.9}{
   \begin{tabular}{cccc}
   \toprule
   Methods & Labeled   & Unlabeled & $AP^{50}$ \\ \hline
   Supervised & \multirow{10}{*}{VOC07} &  \multirow{10}{*}{VOC12} & 76.70 \\
   STAC~\cite{stac} &   &   &  77.45 \\
   Unbiased Teacher v1~\cite{unbiasedv1} &  &  & 77.37 \\
   ISMT~\cite{interactive} &   &   & 77.23 \\
   Instant-Teaching~\cite{instant} &  &  & 79.20 \\
   Humble Teacher~\cite{humble} &  &  & 80.94 \\
   RPL \cite{rethinking} &  &  & 79.00 \\
   ACRST~\cite{ACRST} &  &  & 78.16 \\ 
   Unbiased Teacher v2~\cite{unbiasedv2} &  &  & 81.29 \\
   % \hline
   PCL (Ours) &  &  &  \textbf{82.33} \\
   \bottomrule 
   \end{tabular}}
   \end{center}
   \vspace{-3mm}
 \end{table}
 
%-------------------------------------------------------------------------
\subsection{Noise-unaware learning}
\label{sec:spll} 
After correcting the pseudo-labels, their quality has been improved, but noise may still be present.
Therefore, we propose a noise-unaware strategy to extract crucial knowledge from the noisy pseudo-labels. 
Specifically, we first align the proposals of the student model and the teacher model by an affine transform function, as in ~\cite{softteacher,labelmatch,scmt}.
Then, the proposals of the student are fed into the teacher model to obtain soft outputs, including boxes and scores.
After that, we can calculate the unsupervised loss between the prediction of the student model and the output of the teacher model.

For the unsupervised localization task, we use the corrected boxes as the labels and design a novel re-weighting function.
In general, the smaller the output difference between the student model and the teacher model (i.e., the larger the IoU), the more credible the sample is, so the loss weights should be increased.
However, we get an opposite conclusion: \textbf{designing the loss weight that is negatively correlated with IoU values works better.}
As shown in Fig.~\ref{fig:zhexian1}(d), we conduct a empirical study to analysis the relations between scores and IoU.
It shows that most of the pseudo-labels with high confidence have high IoU values too.
Therefore, we argue that the pseudo-labels, although not as accurate as ground-truth, can also indicate the right direction to optimize the localization task.
When the gap between the predictions and pseudo-labels is large, they are also likely to be far away from the ground-truth. 
Therefore, we design the unsupervised regression loss as follows:
{
\vspace{-2mm}
\begin{equation}
  \label{eq:l-reg}
\begin{aligned}
  \mathcal{L}_{u}^{reg} &=\sum_{i=1}^{n_{pos}} w_i^{reg} L_1(B_i^t, B_i^s), \\
  w_i^{reg} &= sigmoid(\frac{1}{V_{IoU} ^\lambda})
\end{aligned}
\vspace{-2mm}
\end{equation}}where $n_{pos}$ is the number of positive proposals. 
$L_1$ is the MAE loss used in FasterRCNN~\cite{faster}.
$B_i^s$ is the prediction box of the $i$-th proposal from the student model, and $B_i^t$ is the corresponding regression target, i.e., the corrected box.
$V_{IoU}$ is the IoU value between the output boxes of the student and teacher models.
$\lambda $ is a hyper-parameter to control the relative gaps of different IoUs. 

For the unsupervised classification task, we adopt the soft-label technique proposed in LabelMatch~\cite{labelmatch}.
To calculate the unsupervised classification loss, we utilize a soft cross-entropy loss, which is formulated as follows:
{
\vspace{-2mm}
\begin{equation}
  \label{eq:l-cls}
\begin{aligned}
 \mathcal{L}_{u}^{cls} &=\sum_{i=1}^{n_{pro}} w_i^{cls} \sum_{c=1}^C-S_{i, c}^t \log S_{i, c}^s, \\
 w_i^{cls} &= max(S_i^t)
\end{aligned}
\end{equation}}where $n_{pro}$ is the number of proposals, and $C$ is the categories of the dataset. $S_{i, c}^s$ and $S_{i, c}^t$ are the output scores of the $c$-th class and $i$-th proposal from the student and the teacher model, respectively.

\begin{table}[]
   \caption{Ablation study on different components of our method. 
   ``PC'' and ``NL'' indicate pseudo-label correction and noise-unaware learning.
   ``MR'' and ``MW'' denote multi-round refining and multi-vote weighting. 
   $L_{u}^{cls}$ and $L_{u}^{reg}$ are the loss design of our method.}
   \vspace{-3mm}
   \label{table:ablation}
   \begin{center}
   \scalebox{0.9}{
   \begin{tabular}{cccccc}
   \toprule
   \multicolumn{2}{c}{PC}    & \multicolumn{2}{c}{NL}& \multirow{2}{*}{$mAP$} \\
   \cline{1-4}
   MR & MW  & $L_{u}^{cls}$ & $L_{u}^{reg}$  &              \\ \hline
   &  &  &  &  33.8\\
   \hline
   &  & \checkmark&  &  34.1 \\
   &  &\checkmark&\checkmark&34.5 \\
   \checkmark& &\checkmark&\checkmark& 34.7 \\
   \rowcolor{gray!40}\checkmark&\checkmark&\checkmark&\checkmark&35.1 \\
   \bottomrule 
   \end{tabular}}
   \end{center}
   \vspace{-4mm}
 \end{table}
 
 \begin{table}[]
   \caption{Different label and weight settings on $L_{u}^{reg}$. 
   IoU indicates that using IoU value as the loss weight of $L_{u}^{reg}$.
   Inv IoU means that using the loss weight in Eq.~\ref{eq:l-reg}.}
   \vspace{-3mm}
   \label{table:unsupreg}
   \begin{center}
   \scalebox{0.9}{
   \begin{tabular}{ccccc}
   \toprule
   \multicolumn{2}{c}{Loss label}    & \multicolumn{2}{c}{Loss weight}& \multirow{2}{*}{$mAP$} \\
   \cline{1-4}
   Soft & Hard  & IoU & Inv IoU &             \\ \hline
   \checkmark&  &  &  &  33.2\\
   &  \checkmark& &  &  34.1 \\
   &  \checkmark& \checkmark& & 34.2 \\
   \rowcolor{gray!40}&  \checkmark& & \checkmark&34.5 \\
   \bottomrule 
   \end{tabular}}
   \end{center}
   \vspace{-4mm}
 \end{table}

\section{Experiments}
\label{sec:experiments}

\subsection{Dataset and Evaluation}
Our experiments are conducted on MS COCO \cite{microsoft} and PASCAL VOC~\cite{pascal} benchmarks.
MS COCO dataset is split into train2017, unlabeled2017, and val2017.
PASCAL VOC dataset is split into VOC2007-trainval, VOC2012-trainval, and VOC2007-test. 
We follow the recent SSOD methods~\cite{unbiasedv1,labelmatch,unbiasedv2} and use three experimental settings:
(1) \textit{COCO-strandard}: we randomly sample 1\%, 5\% and 10\% ratios of the train2017 as labeled data, and treat the remainder as unlabeled data.
(2) \textit{COCO-additional}: we use all the train2017 as labeled data and the unlabeled2017 as unlabeled data.
The val2017 is used to evaluate the performance on MS COCO benchmark with the metric $mAP$.
(3) \textit{VOC}: we use the VOC2007-trainval as labeled data and the  VOC2012-trainval as unlabeled data.
The VOC2007-test is used to evaluate the performance on PASCAL VOC benchmark with the metric $AP^{50}$.

\subsection{Implementation Details}
We employ FasterRCNN~\cite{faster} with ResNet-50 backbone~\cite{resnet} and FPN~\cite{fpn} as the default detector in our experiments.
However, our method can easily be adapted to other detectors as well.
The pseudo-label selection threshold is set to 0.7, which is the same as SCMT~\cite{scmt}.
The SSOD model is trained on 4 GPUs with 10 images per GPU (2 labeled and 8 unlabeled images) under \textit{COCO-standard} and \textit{VOC}, and 8 images per GPU (4 labeled and 4 unlabeled images) under \textit{COCO-additional}.
The initial learning rate is set to 0.01. 
For \textit{COCO-standard}, the training iteration is 180k, and the learning rate is divided by 10 at 120k and 160k iterations.
For \textit{COCO-additional}, the training iteration is 360k, and the learning rate is divided by 10 at 240k and 320k iterations.
For \textit{VOC}, the training iteration is 60k, and the learning rate is divided by 10 at 40k iterations.
In ablation studies, we train the model with 90k iterations at 10\% labeled ratio to save computational cost.
The data augmentation of this work is followed as SoftTeacher~\cite{softteacher}.
The $n_b$ and $\sigma$ in box jitter function are set 10 and 0.06.
The number of rounds of multi-round refining ($n_r$) is set to 2.
More training and implementation details are presented in the Appendix.

\subsection{Comparison with State-of-the-Art Methods}
The baseline of our method is SoftTeacher~\cite{softteacher}, where we add the pseudo-label correction and learning strategies and achieve significant performance improvements.
Under \textit{COCO-standard}, as shown in Table~\ref{table:mscoco}, our method outperforms SoftTeacher by 3.90,  2.54, and 2.43 mAP, and boosts the accuracy of supervised learning by 12.16, 12.11, and 9.57 mAP. 
The most similar method compared with ours is SCMT~\cite{scmt}, which filters pseudo-labels with a low threshold and utilizes a soft-weight technique to the loss calculation. 
Our proposed PCL shows consistent accuracy improvements at all settings of the MS COCO, with AP values exceeding 1 point.
When compared with other recent SOTA methods, although our method is not the best at the 1\% setting, it outperforms the other methods significantly at 5\% and 10\% settings.
Under \textit{COCO-additional}, as shown in Table~\ref{table:mscoco_additional}, PCL improves the supervised baseline from 37.63 mAP to 44.33 mAP, and also outperforms SoftTeacher~\cite{softteacher} and SCMT~\cite{scmt} by 2.93 mAP and 1.77 mAP, respectively.
Under \textit{VOC}, as illustrated in Table~\ref{table:voc}, PCL achieves 82.33 mAP on $AP^{50}$, pushing the recent state-of-the-arts by a further 1.04 mAP.

\subsection{Ablation study}

\begin{figure}[]
   \begin{center}
   \includegraphics[width=0.8\linewidth]{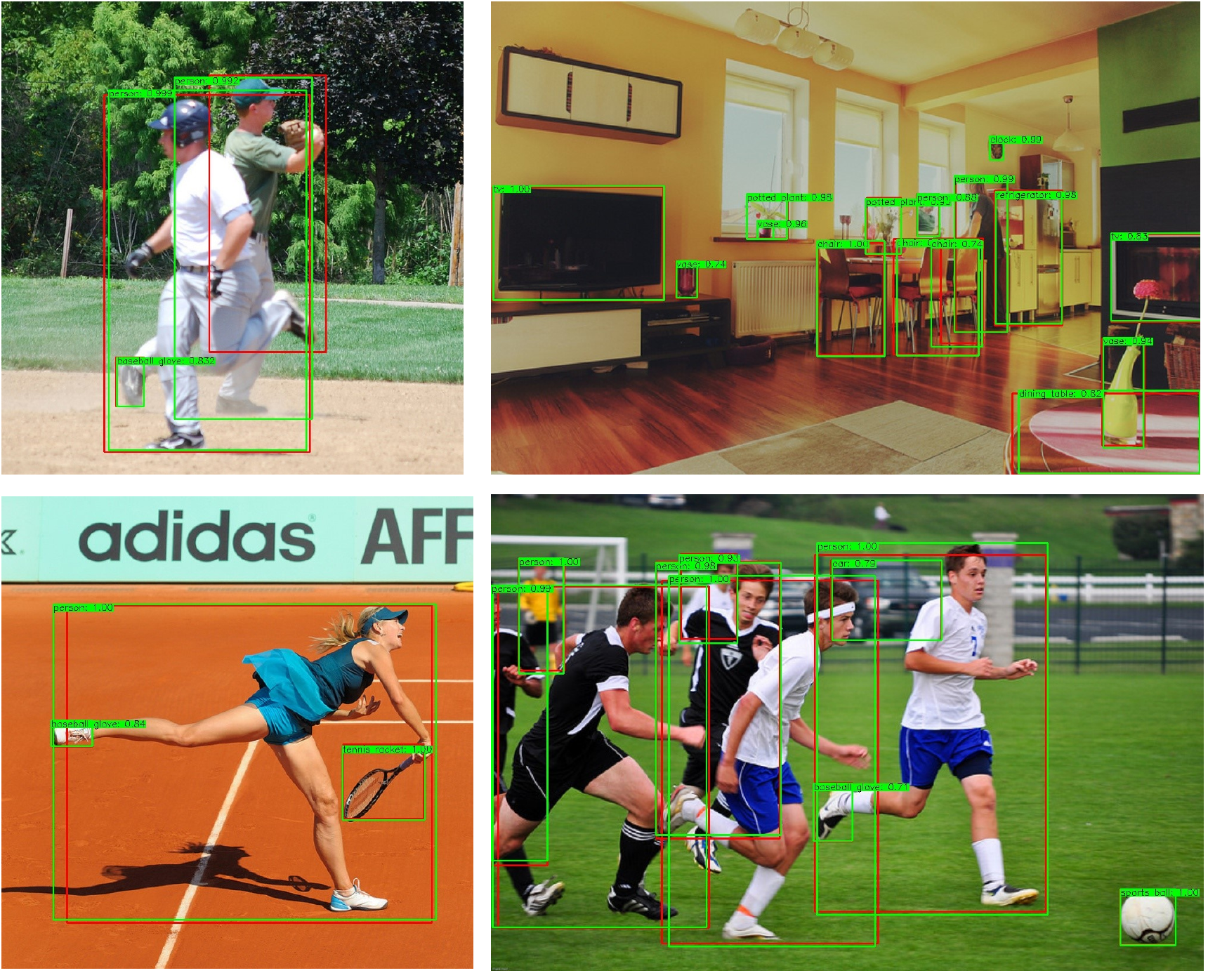}
   \end{center}
   \vspace{-4mm}
   \caption{Visualisation of pseudo-label correction. 
           Red bounding boxes are original pseudo boxes.
           Green bounding boxes are corrected pseudo boxes.}
   \vspace{-4mm}
   \label{fig:show}
\end{figure}

\textbf{Effect of different components:} We conduct ablation experiments at the 10\% labeled data setting to validate the effectiveness of different components in PCL.
As shown in Table~\ref{table:ablation}, after applying the noise-unaware learning strategy, i.e., $L_{u}^{cls}$ in Eq.~\ref{eq:l-cls} and $L_{u}^{reg}$ in Eq.~\ref{eq:l-reg}, mAP is increased from 33.8 to 34.5 mAP.
Adding the pseudo-label correction strategy, i.e., MR (multi-round refining) and MW (multi-vote weighting), further increases mAP from 34.5 to 35.1.
Furthermore, We conduct a visualisation analysis on pseudo-label correction, as shown in Fig.~\ref{fig:show}.
It clearly shows that the position of some inaccurate boxes (drawn in red) has been corrected (drawn in green).
This phenomenon is also consistent with the results of our statistical analysis in Fig.~\ref{fig:zhexian1}(b)(c), proving the effectiveness of our method.

\begin{figure}[]
  \begin{center}
  \includegraphics[width=0.8\linewidth]{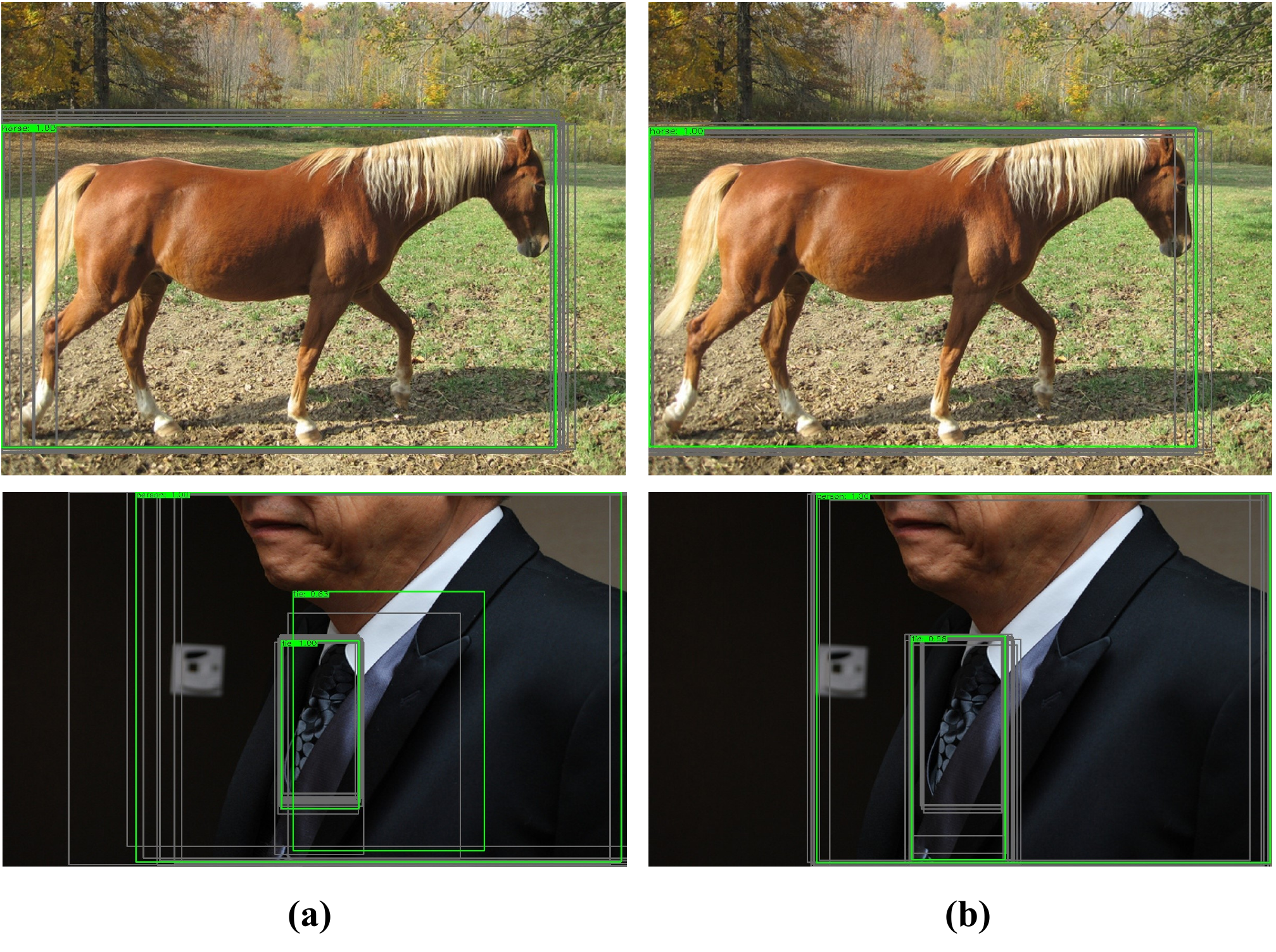}
  \end{center}
  \vspace{-4mm}
  \caption{Visualisation of the prediction boxes with and without NMS. 
  The green boxes and the grey boxes are prediction boxes after and before NMS, respectively.}
  \label{fig:weight}
  \vspace{-4mm}
\end{figure}

\textbf{Effect of unsupervised regression loss:} 
We research the effect of different label and weight settings in unsupervised regression loss.
As shown in Table~\ref{table:unsupreg}, using the hard label and soft weight that is negatively correlated to IoU works the best.
This result suggests that it is important to enhance the learning of prediction boxes that are far from the pseudo boxes.
In addition, using the soft label is not ideal, with an AP gap of 0.9 compared to the hard label.
We guess that not all soft labels generated by the teacher model are accurate, learning them all will make the model confused in training.
Moreover, we conduct a visualisation analyse on the prediction boxes with and without NMS as shown in Fig. ~\ref{fig:weight}. 
It is clear that our PCL has denser prediction boxes than SoftTeacher~\cite{softteacher}.
The design of the loss weight in Eq.~\ref{eq:l-reg} pulls the prediction boxes tightly around the pseudo boxes, avoiding the false-positive detections shown in the bottom case of Fig.~\ref{fig:weight}.

\subsection{Transferability Validation}
In this section, we validate the transferability of our method to other SSOD methods.
Specifically, we have re-implemented the Unbiased Teacher~\cite{unbiasedv1} within our framework and applied the same data augmentation as in ~\cite{softteacher}.
By incorporating the pseudo-label correction (PLC) discussed in Sec~\ref{sec:splc} and the unsupervised regression loss (URL) defined in Eq.~\ref{eq:l-reg}, we have observed a notable improvement of 1.1 mAP in accuracy.
As for SoftTeacher~\cite{softteacher}, we adopted a low threshold towards pseudo-label selection, similar to SCMT~\cite{scmt}, i.e., 0.7.
This modification can significantly alleviate the missing detection problem in SoftTeacher, improving the accuracy from 31.9 mAP to 33.8 mAP.
When we incorporated PLC and URL into SoftTeacher, we observed an additional increase in accuracy of 1.0 mAP. 
In addition, when we applied PLC into PseCo~\cite{pseco}, a more recent SSOD method, there is also a 1.1 mAP improvement.
The consistent improvements in performance demonstrate the general applicability of our method. 
Importantly, since PCL is orthogonal to most existing SSOD methods, it can be easily incorporated to remarkably improve accuracy.

\subsection{Hyper-parameter Analysis}
We have conducted experiments to research the effect of hyper-parameters in our method.
Table~\ref{table:lamda} shows the effect of different $\lambda$ values in $L_{u}^{reg}$.
When $\lambda=5$, we get the optimal result.
Nevertheless, the other selections of $\lambda$ can also get close accuracy, which shows $\lambda$ is not parameter-sensitive.
Table~\ref{table:variance} studies the effect of different variance values in the box jitter function.
When $\sigma=0.06$, The accuracy is the best.
However, when we increase the $\sigma$ to 0.3, the accuracy drops obviously.
This is mainly due to the large box jitter variance causes the prediction boxes to be severely offset from the objects, thus preventing the pseudo boxes from being corrected. 
We set the number of multi-round refining $n_r =2 $ by analysing the differences in the output of multiple rounds.
As shown in Fig.~\ref{fig:zhexian1} (a), it can be clearly observed that after two rounds of refinement, the differences tend to be small.

\begin{table}[t]
  \renewcommand\arraystretch{1.3}
  \caption{Transform effect of our method, i.e., pseudo-label correction (PLC) and unsupervised regression loss (URL).}
  \vspace{-3mm}
  \label{table:unb}
  \begin{center}
  \scalebox{0.9}{
  \begin{tabular}{ccc}
  \toprule
  Method & Strategy   & $mAP$ \\ \hline
  \multirow{2}{*}{Unbiased Teacher~\cite{unbiasedv1}}  &   Baseline    & 33.1 \\
   &   PCL \& URL &  34.2 (+1.1) \\
  \multirow{2}{*}{SoftTeacher~\cite{softteacher}}  &   Baseline    & 33.8 \\
   &   PCL \& URL &  34.8 (+1.0) \\
   \multirow{2}{*}{PseCo~\cite{pseco}}  &   Baseline    & 34.2 \\
   &   PCL &  35.3 (+1.1) \\
  \bottomrule 
  \end{tabular}}
  \end{center}
  \vspace{-3mm}
\end{table}

\begin{table}[]
  \caption{The effect of $\lambda$ in $L_{u}^{reg}$}
  \vspace{-4mm}
  \label{table:lamda}
  \begin{center}
  \scalebox{0.9}{
  \begin{tabular}{ccccc}
  \toprule
  $\lambda$ & 1  & 3 &  5 &  7           \\ \hline
  AP & 34.4 & 34.3 & \textbf{34.5} &  34.3\\
  \bottomrule 
  \end{tabular}}
  \end{center}
  \vspace{-4mm}
\end{table}

\begin{table}[]
  \caption{The effect of different variances in Eq.\ref{eq:jitter}.}
  \vspace{-4mm}
  \label{table:variance}
  \begin{center}
  \scalebox{0.9}{
  \begin{tabular}{cccccc}
  \toprule
  $\sigma$ &0.03 & 0.06  & 0.1 &  0.15 &  0.30 \\ \hline
  AP & 34.8 & \textbf{35.1} & 34.9 & 34.7 &  33.5  \\
  \bottomrule 
  \end{tabular}}
  \end{center}
  \vspace{-4mm}
\end{table}

\section{Conclusion}
\label{sec:conclusion}
In this paper, we present a novel approach to address the issue of localization noise in semi-supervised object detection.
Our proposed pseudo-label correction method is a plug-and-play solution that significantly improves the stability and localization accuracy of pseudo-labels, leading to a better overall performance.
Moreover, we investigate the challenging problem of learning the localization knowledge from noisy pseudo-labels, and introduce a novel regression loss weight design that facilitates precise object localization. 
Our approach is extensively evaluated on popular benchmarks including MS COCO and PASCAL VOC, and the results demonstrate its remarkable effectiveness.

%%%%%%%%% REFERENCES
{\small
\bibliographystyle{ieee_fullname}
\bibliography{egbib}

\begin{thebibliography}{10}\itemsep=-1pt

\bibitem{bachman2014learning}
Philip Bachman, Ouais Alsharif, and Doina Precup.
\newblock Learning with pseudo-ensembles.
\newblock {\em NeurIPS}, 27:3365--3373, 2014.

\bibitem{remixmatch}
David Berthelot, Nicholas Carlini, Ekin~D Cubuk, Alex Kurakin, Kihyuk Sohn, Han
  Zhang, and Colin Raffel.
\newblock Remixmatch: Semi-supervised learning with distribution matching and
  augmentation anchoring.
\newblock In {\em ICLR}, 2019.

\bibitem{mixmatch}
David Berthelot, Nicholas Carlini, Ian Goodfellow, Nicolas Papernot, Avital
  Oliver, and Colin~A Raffel.
\newblock Mixmatch: A holistic approach to semi-supervised learning.
\newblock {\em NeurIPS}, 32, 2019.

\bibitem{labelmatch}
Binbin Chen, Weijie Chen, Shicai Yang, Yunyi Xuan, Jie Song, Di Xie, Shiliang
  Pu, Mingli Song, and Yueting Zhuang.
\newblock Label matching semi-supervised object detection.
\newblock In {\em CVPR}, pages 14381--14390, 2022.

\bibitem{pascal}
Mark Everingham, Luc Van~Gool, Christopher~KI Williams, John Winn, and Andrew
  Zisserman.
\newblock The pascal visual object classes (voc) challenge.
\newblock {\em International journal of computer vision}, 88(2):303--338, 2010.

\bibitem{resnet}
Kaiming He, Xiangyu Zhang, Shaoqing Ren, and Jian Sun.
\newblock Deep residual learning for image recognition.
\newblock In {\em CVPR}, pages 770--778, 2016.

\bibitem{consistency}
Jisoo Jeong, Seungeui Lee, Jeesoo Kim, and Nojun Kwak.
\newblock Consistency-based semi-supervised learning for object detection.
\newblock {\em NeurIPS}, 32, 2019.

\bibitem{interpolation}
Jisoo Jeong, Vikas Verma, Minsung Hyun, Juho Kannala, and Nojun Kwak.
\newblock Interpolation-based semi-supervised learning for object detection.
\newblock In {\em CVPR}, pages 11602--11611, 2021.

\bibitem{laine2016temporal}
Samuli Laine and Timo Aila.
\newblock Temporal ensembling for semi-supervised learning.
\newblock {\em arXiv preprint arXiv:1610.02242}, 2016.

\bibitem{lee2013pseudo}
Dong-Hyun Lee et~al.
\newblock Pseudo-label: The simple and efficient semi-supervised learning
  method for deep neural networks.
\newblock {\em ICML}, page 896, 2013.

\bibitem{MAGCP}
Aoxue Li, Peng Yuan, and Zhenguo Li.
\newblock Semi-supervised object detection via multi-instance alignment with
  global class prototypes.
\newblock In {\em CVPR}, pages 9809--9818, 2022.

\bibitem{pseco}
Gang Li, Xiang Li, Yujie Wang, Shanshan Zhang, Yichao Wu, and Ding Liang.
\newblock Pseco: Pseudo labeling and consistency training for semi-supervised
  object detection.
\newblock In {\em ECCV}, 2022.

\bibitem{rethinking}
Hengduo Li, Zuxuan Wu, Abhinav Shrivastava, and Larry~S Davis.
\newblock Rethinking pseudo labels for semi-supervised object detection.
\newblock In {\em AAAI}, pages 1314--1322, 2022.

\bibitem{fpn}
Tsung-Yi Lin, Piotr Doll{\'a}r, Ross Girshick, Kaiming He, Bharath Hariharan,
  and Serge Belongie.
\newblock Feature pyramid networks for object detection.
\newblock In {\em CVPR}, pages 2117--2125, 2017.

\bibitem{microsoft}
Tsung-Yi Lin, Michael Maire, Serge Belongie, James Hays, Pietro Perona, Deva
  Ramanan, Piotr Doll{\'a}r, and C~Lawrence Zitnick.
\newblock Microsoft coco: Common objects in context.
\newblock In {\em ECCV}, pages 740--755. Springer, 2014.

\bibitem{unbiasedv1}
Yen-Cheng Liu, Chih-Yao Ma, Zijian He, Chia-Wen Kuo, Kan Chen, Peizhao Zhang,
  Bichen Wu, Zsolt Kira, and Peter Vajda.
\newblock Unbiased teacher for semi-supervised object detection.
\newblock In {\em ICLR}, 2021.

\bibitem{unbiasedv2}
Yen-Cheng Liu, Chih-Yao Ma, and Zsolt Kira.
\newblock Unbiased teacher v2: Semi-supervised object detection for anchor-free
  and anchor-based detectors.
\newblock In {\em CVPR}, pages 9819--9828, 2022.

\bibitem{miyato2018virtual}
Takeru Miyato, Shin-ichi Maeda, Masanori Koyama, and Shin Ishii.
\newblock Virtual adversarial training: a regularization method for supervised
  and semi-supervised learning.
\newblock {\em IEEE TPAMI}, 41(8):1979--1993, 2018.

\bibitem{faster}
Shaoqing Ren, Kaiming He, Ross Girshick, and Jian Sun.
\newblock Faster r-cnn: Towards real-time object detection with region proposal
  networks.
\newblock {\em NeurIPS}, 28, 2015.

\bibitem{sajjadi2016regularization}
Mehdi Sajjadi, Mehran Javanmardi, and Tolga Tasdizen.
\newblock Regularization with stochastic transformations and perturbations for
  deep semi-supervised learning.
\newblock {\em NeurIPS}, 29:1163--1171, 2016.

\bibitem{fixmatch}
Kihyuk Sohn, David Berthelot, Nicholas Carlini, Zizhao Zhang, Han Zhang,
  Colin~A Raffel, Ekin~Dogus Cubuk, Alexey Kurakin, and Chun-Liang Li.
\newblock Fixmatch: Simplifying semi-supervised learning with consistency and
  confidence.
\newblock {\em NeurIPS}, 33:596--608, 2020.

\bibitem{stac}
Kihyuk Sohn, Zizhao Zhang, Chun-Liang Li, Han Zhang, Chen-Yu Lee, and Tomas
  Pfister.
\newblock A simple semi-supervised learning framework for object detection.
\newblock {\em arXiv preprint arXiv:2005.04757}, 2020.

\bibitem{humble}
Yihe Tang, Weifeng Chen, Yijun Luo, and Yuting Zhang.
\newblock Humble teachers teach better students for semi-supervised object
  detection.
\newblock In {\em CVPR}, pages 3132--3141, 2021.

\bibitem{meanteacher}
Antti Tarvainen and Harri Valpola.
\newblock Mean teachers are better role models: Weight-averaged consistency
  targets improve semi-supervised deep learning results.
\newblock {\em NeurIPS}, 30, 2017.

\bibitem{scmt}
Feng Xiong, Jiayi Tian, Zhihui Hao, Yulin He, and Xiaofeng Ren.
\newblock Scmt: Self-correction mean teacher for semi-supervised object
  detection.
\newblock In {\em IJCAI}, pages 1488--1494, 2022.

\bibitem{softteacher}
Mengde Xu, Zheng Zhang, Han Hu, Jianfeng Wang, Lijuan Wang, Fangyun Wei, Xiang
  Bai, and Zicheng Liu.
\newblock End-to-end semi-supervised object detection with soft teacher.
\newblock In {\em CVPR}, pages 3060--3069, 2021.

\bibitem{interactive}
Qize Yang, Xihan Wei, Biao Wang, Xian-Sheng Hua, and Lei Zhang.
\newblock Interactive self-training with mean teachers for semi-supervised
  object detection.
\newblock In {\em CVPR}, pages 5941--5950, 2021.

\bibitem{ACRST}
Fangyuan Zhang, Tianxiang Pan, and Bin Wang.
\newblock Semi-supervised object detection with adaptive class-rebalancing
  self-training.
\newblock In {\em AAAI}, pages 3252--3261, 2022.

\bibitem{simmatch}
Mingkai Zheng, Shan You, Lang Huang, Fei Wang, Chen Qian, and Chang Xu.
\newblock Simmatch: Semi-supervised learning with similarity matching.
\newblock In {\em CVPR}, pages 14471--14481, 2022.

\bibitem{zheng2022simmatch}
Mingkai Zheng, Shan You, Lang Huang, Fei Wang, Chen Qian, and Chang Xu.
\newblock Simmatch: Semi-supervised learning with similarity matching.
\newblock {\em CVPR}, pages 14471--14481, 2022.

\bibitem{dense}
Hongyu Zhou, Zheng Ge, Songtao Liu, Weixin Mao, Zeming Li, Haiyan Yu, and Jian
  Sun.
\newblock Dense teacher: Dense pseudo-labels for semi-supervised object
  detection.
\newblock In {\em ECCV}, pages 35--50. Springer, 2022.

\bibitem{instant}
Qiang Zhou, Chaohui Yu, Zhibin Wang, Qi Qian, and Hao Li.
\newblock Instant-teaching: An end-to-end semi-supervised object detection
  framework.
\newblock In {\em CVPR}, pages 4081--4090, 2021.

\end{thebibliography}
}

\end{document}